\documentclass[10pt,twocolumn,letterpaper]{article}

\usepackage[pagenumbers]{cvpr} 
\usepackage[
  left=0.8in,
  right=0.8in,
  top=0.8in,
  bottom=0.8in
]{geometry}

\usepackage{graphicx}

\usepackage[pagebackref,breaklinks,colorlinks,allcolors=cvprblue]{hyperref}
\usepackage{tikz}
\usepackage{adjustbox}
\usepackage{multirow}
\usepackage[table,xcdraw,dvipsnames]{xcolor}
\usepackage{tcolorbox}
\usepackage{graphicx}
\usepackage{booktabs}
\usepackage{float}
\usepackage{colortbl}
\usepackage{amsmath}
\usepackage{makecell}

\definecolor{datasetcolor}{HTML}{4472C4} 
\usepackage{tikz}
\definecolor{cvprblue}{rgb}{0.21,0.49,0.74}
\definecolor{firstb}{HTML}{FF6767}
\definecolor{thirdb}{HTML}{3564FF}
\definecolor{fovgreen}{HTML}{BED8CB}

\usepackage[pagebackref,breaklinks,colorlinks,allcolors=cvprblue]{hyperref}

\title{%
Spatial Calibration of Diffuse LiDARs
\vspace{-12pt}
}

\author{
Nikhil Behari\\
MIT\\
{\tt\small nbehari@mit.edu}
\and
Luke Apostolides \\
MIT\\
{\tt\small lapost3@mit.edu}
\and
Ramesh Raskar\\
MIT\\
{\tt\small raskar@mit.edu}
}

\begin{document}

\twocolumn[{%
\renewcommand\twocolumn[1][]{#1}%
\maketitle
\begin{center}
    \centering
    \captionsetup{type=figure}
    \vspace{-18pt}
    \includegraphics[width=\textwidth]{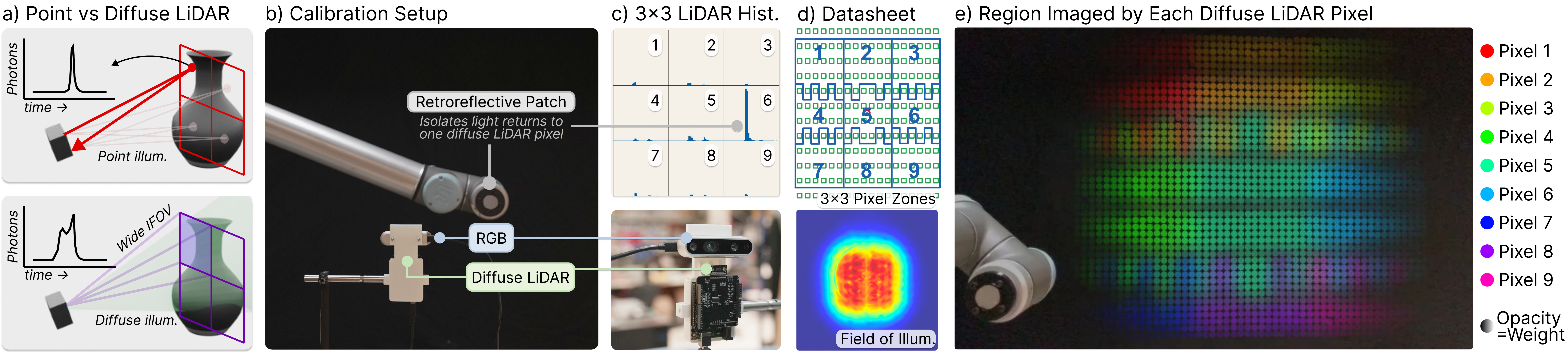} 
    \captionof{figure}{\textbf{Diffuse LiDAR spatial calibration overview}. Diffuse LiDARs, used in consumer devices and mobile robots, (a) aggregate time-of-flight returns over wide per-pixel fields-of-view, producing spatially mixed measurements that make standard LiDAR-to-RGB calibration difficult. We describe (b-d) a simple method that calibrates diffuse LiDAR pixels by estimating each pixel’s effective support region and spatial weighting over a co-located RGB image plane. The (e) resulting per-pixel response maps provide an explicit LiDAR-to-RGB correspondence for accurate cross-modal alignment and fusion.}
    \label{fig:teaser}
\end{center}%
}]

\maketitle

\noindent \textbf{tl;dr} Diffuse LiDARs integrate depth returns over a wide pixel field-of-view, making standard calibration with RGB cameras difficult; we describe a simple method for estimating each diffuse LiDAR pixel’s footprint and relative spatial sensitivity in a co-located RGB image plane. 

\section{Background}
\label{sec:intro}

Light detection and ranging (LiDAR) is often paired with RGB imaging to improve depth estimation for 3D reconstruction and 3D perception. In direct time-of-flight (DToF) LiDAR systems (hereafter \textit{LiDAR}), depth is estimated by emitting a short laser pulse and measuring photon return times; photon detections are then timestamped and accumulated into per-pixel histograms of arrival times. Depth is typically estimated from the time bin associated with the dominant histogram peak (e.g., via peak fitting).

Many LiDAR-to-camera calibration methods assume each LiDAR pixel corresponds to a single, well-defined scene point that can be projected into the camera image. This approximation is reasonable for conventional LiDARs (\cref{fig:teaser}a) with narrow per-pixel angular support, where each pixel measures returns from a small angular cone. Under this assumption, LiDAR-to-camera calibration can be done by aligning shared scene features (e.g., planes, edges, reflective targets) since each LiDAR pixel subtends a small solid angle and can be approximated as a single ray that yields a camera-projectable 3D point. 

\textit{Diffuse LiDAR} refers to a class of dToF sensors that differ from conventional LiDARs in two ways \cite{behari2025blurred,sifferman2025recovering,mu2024towards}. First, they use flood illumination rather than a narrow-beam (or collimated) laser transmitter. Second, each reported pixel \textit{aggregates} photon detections over a large instantaneous field-of-view (IFOV), mixing contributions from different regions of the scene within each histogram measurement. As a result, each reported LiDAR pixel encodes \textit{spatially mixed} depth returns rather than a single-point depth.

This LiDAR design introduces unique trade-offs. The large instantaneous field-of-view and coarse effective resolution make dense, high-fidelity 3D reconstruction challenging; these sensors are therefore primarily used for coarse proximity sensing in mobile and robotic settings. At the same time, their low cost (i.e., below \$10 \cite{ams_osram_tmf882x,st_VL53L8CX}) and small form factor have enabled LiDAR-guided perception in resource-constrained platforms, including applications in navigation and planning \cite{young_enhancing_2025,sifferman2025efficient}, scene and material understanding \cite{callenberg2021low,behari2025blurred,sifferman2025recovering}, and non-line-of-sight imaging \cite{denali2026}.

A central challenge is integrating diffuse LiDAR with other modalities such as RGB. Because each reported pixel mixes returns over a wide angular cone, the measurements do not correspond to a single, well-defined camera/LiDAR ray, violating assumptions used by standard intrinsic and extrinsic calibration procedures. This can limit cross-modal alignment, fusion, and downstream reconstruction unless the sensor’s wide IFOV response is explicitly modeled.

Below, we present a simple spatial calibration procedure for \textit{diffuse} LiDARs. The method (i) estimates each LiDAR pixel’s footprint (\textit{effective support region}) in the co-located RGB image plane, and (ii) estimates a normalized spatial sensitivity within that footprint, capturing how flood illumination and pixel aggregation weight returns across the region. Unlike approaches that require an external active illumination source (e.g., \cite{callenberg2021low}), we calibrate using only a passive retroreflective target. Our method provides an explicit LiDAR-to-RGB correspondence that can support multimodal alignment and fusion for this sensor class.

\section{Calibration Method}
\label{sec:cal}

Diffuse LiDAR pixels integrate photon returns over a wide instantaneous field-of-view, so a single LiDAR pixel does not correspond to a single point or ray in the RGB image. We estimate, for each LiDAR pixel $p$, a response map in RGB pixel coordinates that (i) identifies the pixel's effective support region in the RGB image and (ii) provides relative spatial weights within that region. These maps support LiDAR-RGB alignment and cross-modal fusion.

Our full calibration resources (sensor mounts, capture/processing scripts, and example outputs) are available at
\href{https://github.com/nikhilbehari/diffuselidar_calibrate}{github.com/nikhilbehari/diffuselidar\_calibrate}.

\subsection{Hardware and Rigid Mount}
\begin{figure}[h]
  \centering
  \includegraphics[width=\linewidth]{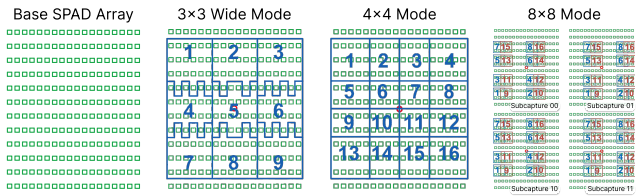}
  \vspace{-18pt}
  \caption{Pixel aggregation modes on the ams OSRAM TMF8828 diffuse dToF LiDAR \cite{ams_osram_tmf882x}. The sensor supports multiple spatial aggregation layouts; in each mode, reported pixels integrate photon returns over a wide instantaneous field-of-view under flood illumination. Our example calibration uses \texttt{3$\times$3 Wide} mode ($P{=}9$).}
  \label{fig:pixel_modes}
  \vspace{-6pt}
\end{figure}

We use an ams OSRAM TMF8828 diffuse dToF module (940\,nm) that reports per-pixel photon-arrival histograms with $T=128$ time bins and supports multiple pixel aggregation modes. We illustrate these modes in \cref{fig:pixel_modes}; in our example calibration below, we operate in \texttt{3$\times$3 Wide} mode ($P=9$). The TMF8828 supports short-range (1.5\,m) and long-range (5\,m) ranging configurations; we perform calibration in both ranging modes to verify that the estimated spatial response is consistent. We capture RGB images with an Intel RealSense D435i at 848$\times$480 resolution. 

\begin{figure}[h]
  \centering
  \includegraphics[width=\linewidth]{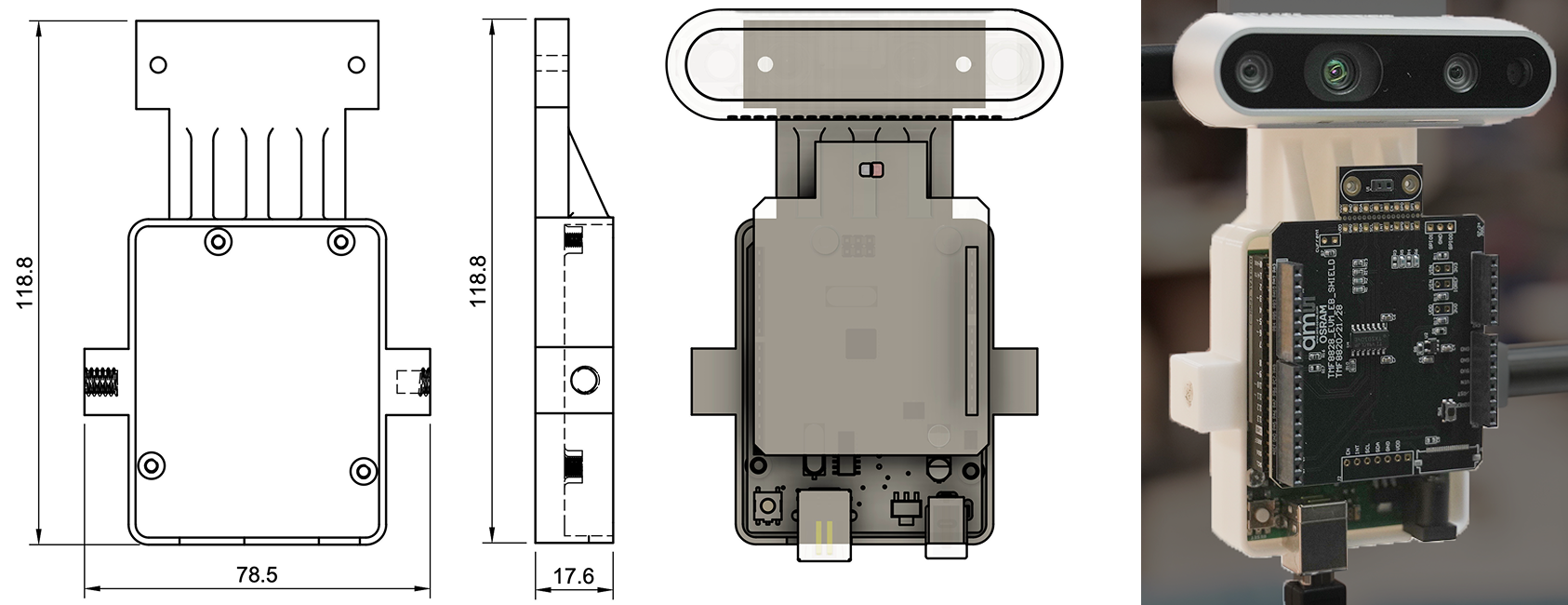}
  \vspace{-18pt}
  \caption{Custom rigid mount for co-located diffuse LiDAR (TMF8828) and RGB (RealSense D435i). Left: CAD mount design (dimensions in mm), showing the fixed sensor-to-sensor geometry and mounting hole layout. Right: real capture mount used for our calibration data collection.}
    
  \label{fig:rigid_mount}
  \vspace{-6pt}
\end{figure}

We rigidly mount the RGB camera and LiDAR on a custom bracket that fixes their relative pose during capture. The bracket places the RGB and LiDAR sensors as close together as possible and aligns their optical axes to be approximately parallel to maximize overlap in the observed field of view. We show our rigid mount design in \cref{fig:rigid_mount}; the full design is available for download on our \href{https://github.com/nikhilbehari/diffuselidar_calibrate}{GitHub page}.

\subsection{Retroreflective Patch Scan}
\begin{figure}[tb]
  \centering
  \includegraphics[width=\linewidth]{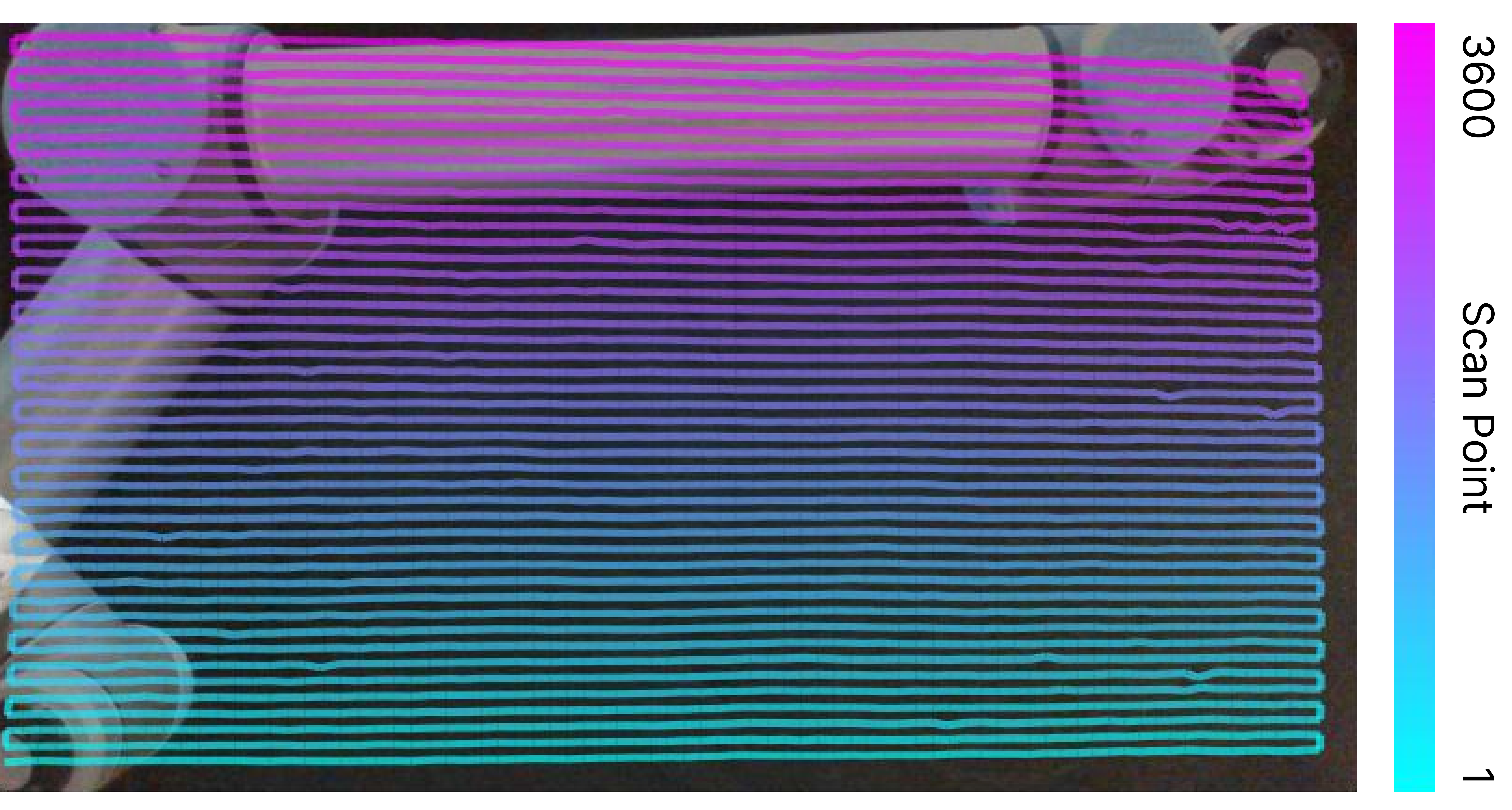}
  \vspace{-16pt}
   \caption{Retroreflective patch scan grid sampled with a UR10 robot arm. We traverse an $80{\times}45$ grid ($K{=}3600$) using a snake pattern to reduce motion between points. At each grid location, we record synchronized RGB frames and per-pixel LiDAR histograms; an identical patch-removed scan is also captured for background subtraction.}
  \label{fig:scan_point}
  \vspace{-6pt}
\end{figure}

We scan a small retroreflective patch across the shared sensor field of views while recording LiDAR histograms and synchronized RGB frames. We use a UR10 robot arm to move a circular retroreflective patch mounted in a 3.4\,mm diameter insert. The patch is oriented approximately fronto-parallel to the RGB camera and moved over a uniform 2D grid at approximately fixed range. In our captures, we traverse an 80$\times$45 grid ($K = 3600$ points) in a snake pattern to reduce robot arm motion between adjacent points, shown in \cref{fig:scan_point}.

To separate out signal from the robot arm and background, we capture two scans with identical robot motion and sensor settings:
(i) a patch-present scan and (ii) a patch-removed scan that serves as the background measurement. We use the second scan only for background subtraction during processing.

\subsection{Histogram Mixing Model}
We index LiDAR pixels by $p\in\{1,\dots,P\}$ and histogram bins by $t\in\{1,\dots,T\}$. At scan index $k\in\{1,\dots,K\}$, the sensor returns a per-pixel photon-arrival histogram $\tau_{p,k}(t)\in\mathbb{N}$. Under wide instantaneous field-of-view operation, each pixel aggregates returns from a spatial \textit{region} rather than a single direction. We model this mixing in the RGB image coordinate system as
\begin{equation}
\tau_{p,k}(t)\;=\;\int_{\Omega} w_p(\mathbf{u})\,\tau_k(\mathbf{u},t)\,d\mathbf{u},
\label{eq:diffuse_hist_mix}
\end{equation}
where $\mathbf{u}\in\mathbb{R}^2$ denotes continuous RGB pixel coordinates over the camera field of view $\Omega$, $\tau_k(\mathbf{u},t)$ is the latent transient response at location $\mathbf{u}$ for capture $k$, and $w_p(\mathbf{u})\ge 0$ is the unknown spatial sensitivity of LiDAR pixel $p$ expressed in RGB coordinates. We estimate $w_p$ over the RGB field of view $\Omega$; any LiDAR sensitivity outside $\Omega$ is unobserved and therefore truncated by our calibration. Our aim in calibration is to estimate $w_p(\mathbf{u})$ up to a per-pixel scale, sampled at scan locations $\{\mathbf{u}_k\}_{k=1}^K$.

\begin{figure*}[tbh]
  \centering
  \includegraphics[width=\textwidth]{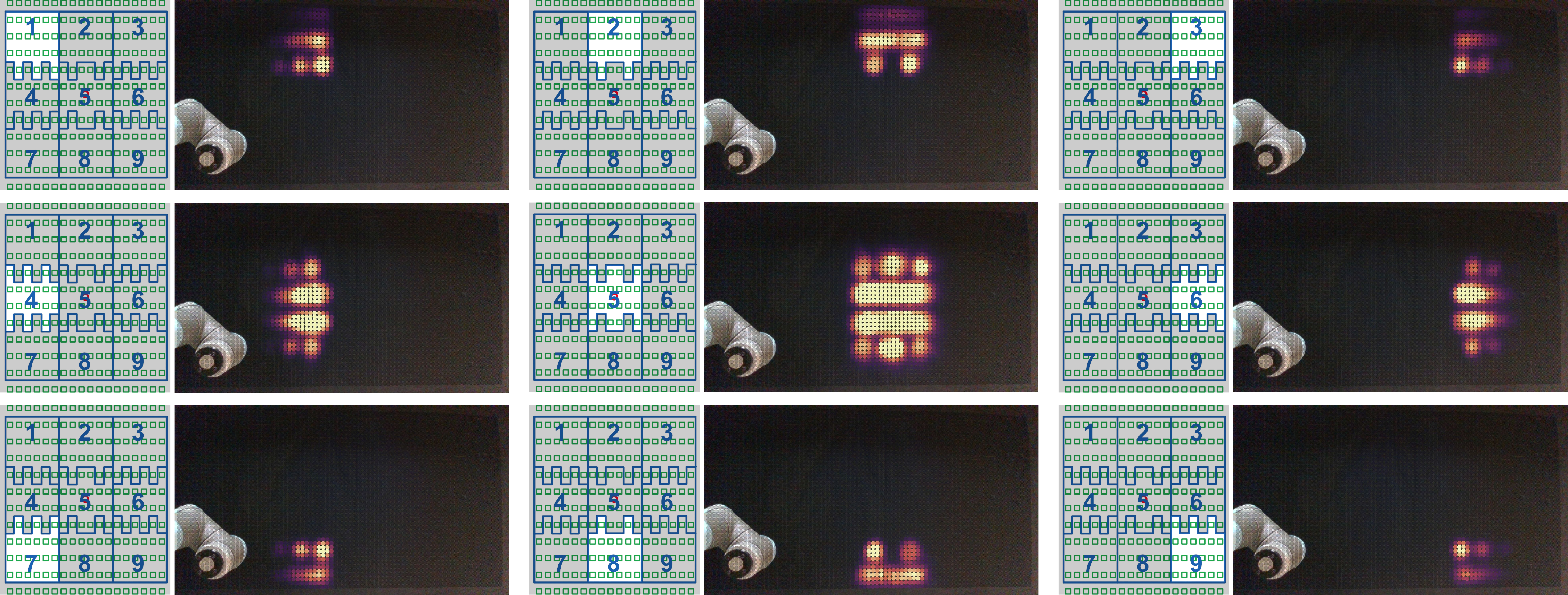}
  \vspace{-16pt}
  \caption{Per-pixel spatial response maps for the TMF8828 \texttt{3$\times$3 Wide} mode overlaid on the co-located RGB image. Nonzero regions show the pixel's effective support in RGB coordinates, and the response magnitudes encode relative spatial sensitivity within that support.}
  \label{fig:per_pixel}
  \vspace{-6pt}
\end{figure*}

\subsection{Processing and Response Map Estimation}
For each scan index $k$, we detect the retroreflective patch in the RGB image using Hough circle detection, yielding per-scan point patch centers $\mathbf{u}_k=(x_k,y_k)$. 

Let $\tau_{p,k}(t)$ denote the histogram for LiDAR pixel $p$ with the patch present, and $\tau^{\mathrm{bg}}_{p,k}(t)$ the corresponding histogram from the patch-removed background scan. We choose a fixed window of histogram-bin indices $G\subset\{1,\dots,T\}$ whose bins correspond to the patch depth. We convert each histogram pair into a scalar patch response
\begin{equation}
R_p(\mathbf{u}_k)\;\triangleq\;\max_{t\in G}\Big[\tau_{p,k}(t)-\tau^{\mathrm{bg}}_{p,k}(t)\Big]_+
\label{eq:patch_response}
\end{equation}

\noindent where $[z]_+ \triangleq \max(z,0)$. We subtract background within $G$, clip negative values to zero, and summarize the windowed return by its maximum photon count. In the sensor's linear (non-saturated) operating regime, the retroreflective patch dominates returns within $G$, so $R_p(\mathbf{u}_k)$ is proportional (up to per-pixel scale and blur due to finite patch size) to the spatial mixing kernel value $w_p(\mathbf{u}_k)$ at the patch location. Thus, the samples $\{(\mathbf{u}_k,R_p(\mathbf{u}_k))\}_{k=1}^K$ provide a response map for pixel $p$ in RGB coordinates.

We store $R_p(\mathbf{u}_k)$ on the scan grid to form a 2D map aligned to the RGB image: its nonzero region estimates the effective support of pixel $p$, and its magnitudes encode relative spatial sensitivity within that support. When useful for downstream fusion, we normalize each pixel map by its peak response:
\begin{equation}
\widetilde{R}_p(\mathbf{u}_k)\;=\;\frac{R_p(\mathbf{u}_k)}{\max_{k'} R_p(\mathbf{u}_{k'})}.
\label{eq:response_norm_max}
\end{equation}

\section{Example Calibration}
We demonstrate the full spatial calibration for the TMF8828 operating in \texttt{3×3 Wide} mode ($P=9$) using the 80×45 scan grid described above. Figure~\ref{fig:teaser} shows the complete set of per-pixel response maps overlaid on the RGB image. Each map represents the sampled mixing kernel $w_p(\mathbf{u})$ up to an unknown per-pixel scale; visualized response maps are peak-normalized per pixel (Eq.~\ref{eq:response_norm_max}) and displayed in RGB coordinates. We additionally visualize isolated per-pixel response maps in Fig.~\ref{fig:per_pixel}, highlighting that the calibration recovers not only each pixel’s support region but also its relative sensitivity variation within that region.

We repeated the calibration in the TMF8828 short-range (1.5m) and long-range (5m) modes to test cross-mode consistency of the recovered response maps. The estimated short and long-range mode response maps agree closely (across 9 pixels: IoU of support masks $=0.915\pm0.029$; centroid displacement $=2.94\pm0.67$ px; cosine similarity between peak-normalized maps $=0.984\pm0.008$). This indicates that the spatial response is largely ranging-mode invariant and that our calibration is repeatable.

Qualitatively, the recovered per-pixel supports match the nominal zone layout reported in the TMF8828 datasheet (Fig.~\ref{fig:teaser}d). Unlike the datasheet, however, our calibration estimates the relative spatial sensitivity within each zone and the aggregate illumination profile across pixels, which we believe can enable more physically grounded LiDAR rendering and LiDAR-RGB fusion. 

\section{Limitations}
Our calibration assumes controlled capture with a rigid mount and a dense retroreflector scan. We estimate per-pixel response maps only in the co-registered RGB image plane; extending these image-plane correspondences to a full 3D (world-space) geometric calibration is possible but outside the scope of this work. Our calibration is recovered at discrete scan locations rather than as a continuous image-plane kernel (which could, however, be fit from our samples). Finally, because spatial weights are estimated using a high-SNR retroreflector, they may not fully capture behavior in real scenes where reflectance and materials vary within a pixel’s effective support.

{
    \small
    \bibliographystyle{ieeenat_fullname}
    \bibliography{main}
}

\end{document}